 \documentclass[a4paper, 10pt, conference]{ieeeconf}      

\IEEEoverridecommandlockouts                    

\usepackage[utf8]{inputenc}
\usepackage[english]{babel}
\usepackage{setspace}
\usepackage{cite}
\usepackage{multirow}
\usepackage[table,xcdraw]{xcolor}
\usepackage{subfigure}
\usepackage{graphics} 
\usepackage{graphicx}
\graphicspath{{figures/}}
\usepackage[mathscr]{euscript}
\usepackage{listings}
\usepackage{color}
\usepackage{listings}

\usepackage{graphicx,wrapfig,lipsum}

\usepackage{tikz}
\usetikzlibrary{intersections}
\usetikzlibrary{decorations.pathreplacing}

\usepackage{multicol}
\usepackage[noend]{algpseudocode}

\usepackage{amsthm} 
\usepackage{amsmath} 
\usepackage{amssymb}  
\usepackage[percent]{overpic}
\usepackage{chemformula}
\usepackage{multirow}
\usepackage{array}

\theoremstyle{plain}

\definecolor{lightgray}{gray}{0.97}
\definecolor{lightgray2}{gray}{0.1}

\title{\LARGE \bf
Proprioceptive Sensing of Soft Tentacles with Model Based Reconstruction for Controller Optimization
}

\author{Andrea Vicari*\textsuperscript{1,2,3}, Nana Obayashi*\textsuperscript{1}, Francesco Stella\textsuperscript{\textdagger 1,4}, Gaetan Raynaud\textsuperscript{\textdagger 5}, \\Karen Mulleners\textsuperscript{5}, Cosimo Della Santina\textsuperscript{4,6}, and Josie Hughes\textsuperscript{1}
\thanks{*Equally contributing first authors. \textdagger Equally contributing second authors.}
\thanks{\textsuperscript{1}CREATE Lab, EPFL, Lausanne, Switzerland. \textsuperscript{2}Scuola Superiore Sant’Anna, Pisa, Italy. \textsuperscript{3}Università di Pisa, Pisa, Italy. \textsuperscript{4}Department of Cognitive Robotics, Delft University of Technology, Delft, The Netherlands. \textsuperscript{5}UNFoLD, EPFL, Lausanne, Switzerland. \textsuperscript{6}Institute of Robotics and Mechatronics, German Aerospace Center (DLR), Wessling, Germany. Contact emails: {\tt\small  andrea.vicari@santannapisa.it, nana.obayashi@epfl.ch, josie.hughes@epfl.ch}.}%
}

\begin{document}

\maketitle
\thispagestyle{empty}
\pagestyle{empty}

\begin{abstract}
The success of soft robots in displaying emergent behaviors is tightly linked to the compliant interaction with the environment. However, to exploit such phenomena, proprioceptive sensing methods which do not hinder their softness are needed. In this work we propose a new sensing approach for soft underwater slender structures based on embedded pressure sensors and use a learning-based pipeline to link the sensor readings to the shape of the soft structure. Using two different modeling techniques, we compare the pose reconstruction accuracy and identify the optimal approach. Using the proprioceptive sensing capabilities we show how this information can be used to assess the swimming performance over a number of metrics, namely swimming thrust, tip deflection, and the traveling wave index. We conclude by demonstrating the robustness of the embedded sensor on a free swimming soft robotic squid swimming at a maximum velocity of 9.5 cm/s, with the absolute tip deflection being predicted within an error less than 9\% without the aid of external sensors.  
\end{abstract}

\section{Introduction}
\label{sec:intro}

Soft underwater animals such as octopuses leverage their softness when interacting with water to achieve complex manipulation actions or to maximize thrust generation~\cite{kim2013soft}.
Proprioceptive sensing---or `self-sensing' of its own body---is particularly key for these animals, as their limbs and structure are inherently underactuated~\cite{wang2018toward}.
To be able to perform sensory-motor control of their limbs and adapt to changing environmental conditions such as currents, they must be able to understand the configuration of the body within a fluid.
This is also true for soft underwater bio-inspired robots which leverage soft underactuated or passive structures to locomote. 
Understanding the body deformation of the soft robot can allow for the optimization of actuation, and also understanding of interactions with the environment.


\begin{figure}[tb]
    \centering
    \includegraphics[width=1\columnwidth]{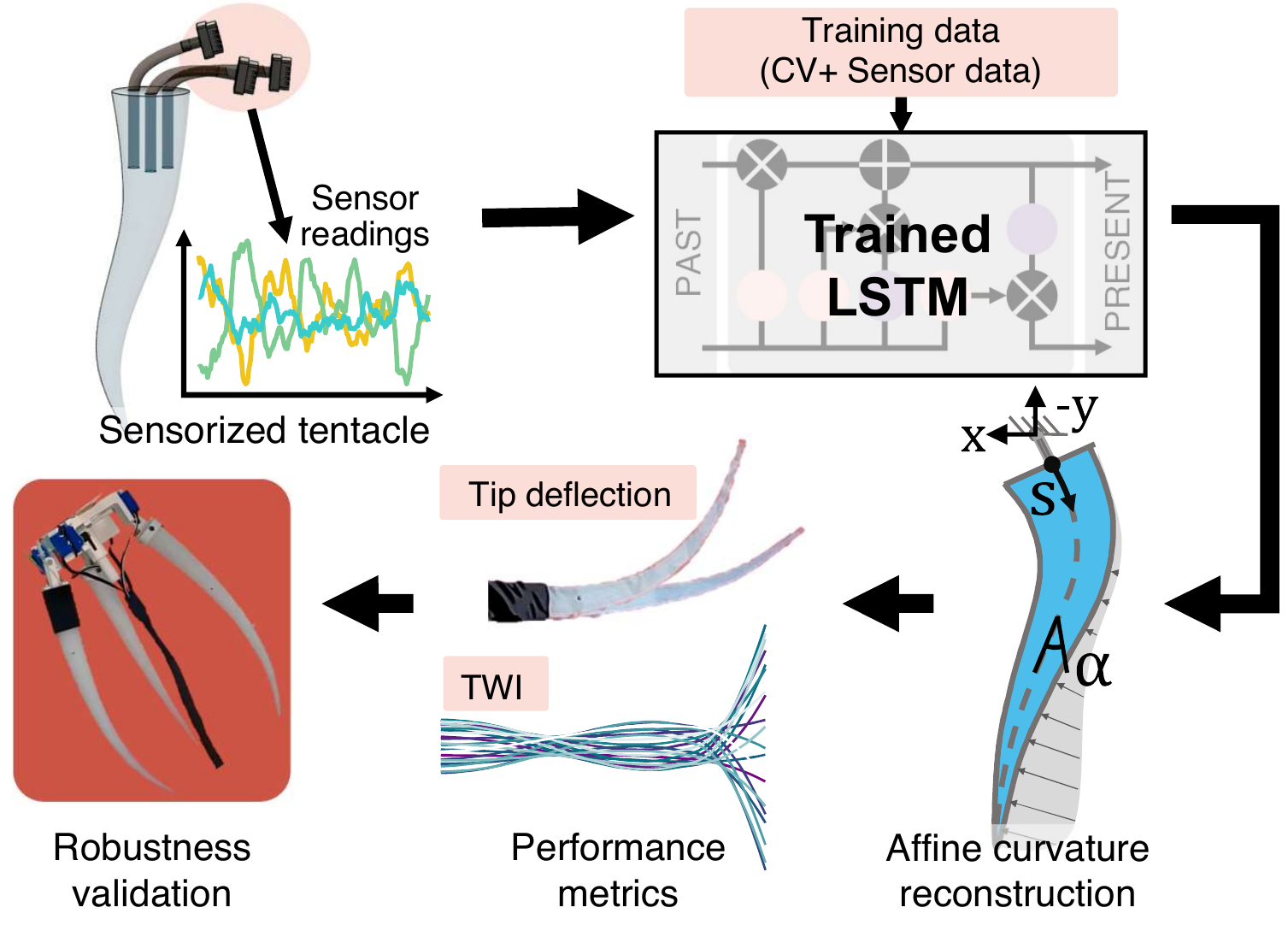}
    \caption{Summary of our approach to use model-based reconstruction from sensor readings to tentacle shape to be utilized for assessing swimming performance metrics. Validation experiments are then performed on a full robot to assess robustness.}
    \label{fig:fig1}
    \vspace{-4mm}
\end{figure}


Building soft robots or structures with distributed sensing capabilities is challenging ~\cite{wang2018toward}.  This is particularly so for underwater swimmers where we can observe a plethora of motions and structure are typically under actuated. To avoid restricting the soft structure's motions by adding rigid parts, these sensors must be compliant and also robust to diverse environmental characteristics. Previous work has addressed this challenge by using soft tentacles as a physical computational reservoir with proprioceptive strain sensors and vision markers but this reduced the compliance of the tentacle~\cite{judd2019sensing}.  Other proprioceptive sensors only detect unidirectional deformation~\cite{xie2020proprioceptive} or are only activated under relatively large environmental disturbances~\cite{wang2021integration}. Several works used recurrent neural networks to predict the dynamics of a soft structure but their methods were not applicable to robust actuation conditions or environments~\cite{thuruthel2019soft,loo2022robust}.

Researchers of the theory of embodied intelligence agree that coupling morphology and environment is crucial for developing sensorimotor loops~\cite{lungarella2006mapping}. We must exploit proprioceptive sensing for control of the body so that it can be utilized to give accurate feedback on performance characteristics in various environments. To date, dynamic parameter estimation of soft underwater tentacles using computer vision and affine curvature fitting~\cite{stella2022experimental} or pose reconstruction of soft underwater hands~\cite{wang2022deep} require laboratory experiments with access to a camera and are unfit for online optimization in real-world environmental conditions. 

In this work, we develop sensorized soft tentacles that have proprioception capabilities, and on understanding how the control affects the thrust generation using proprioceptive sensor values, our approach summarized visually in Fig.~\ref{fig:fig1}. We propose using soft fluidic sensors~\cite{hughes2020sensorization} at the base of the tentacle, where the pressure change at the root can be used to infer the behavior at the lower segments of the tentacle. This allows the tentacle to remain fully soft, not restricted by sensors, and requires only a few sensors to predict the motion.  By using computer vision, we can develop a training dataset that can be used in conjunction with the LSTM (long short-term memory) network and constrain the output to an affine curvature model, to improve the accuracy and better capture the dynamics. 
By developing a soft passive tentacle with embedded sensors, we can use the reconstructed tentacle shape to obtain an indication of performance through a model-based approach. Metrics such as tip deflection and traveling wave index (TWI) are explored as indicators of thrust generation and efficiency.
We demonstrate this approach first on a single tentacle, and then on a free-swimming squid-inspired robot developed in~\cite{obayashi2022soft} to exhibit the accuracy and robustness of the sensor technology and to validate the single tentacle experiments. 


\section{Methods}
In this section, we introduce the design fabrication of the sensor and the reconstruction methods. The metrics for utilizing this proprioceptive reconstruction are then presented.

\subsection{Sensor design \& fabrication}

\begin{figure}[tb]
    \centering
    \includegraphics[width=1\columnwidth]{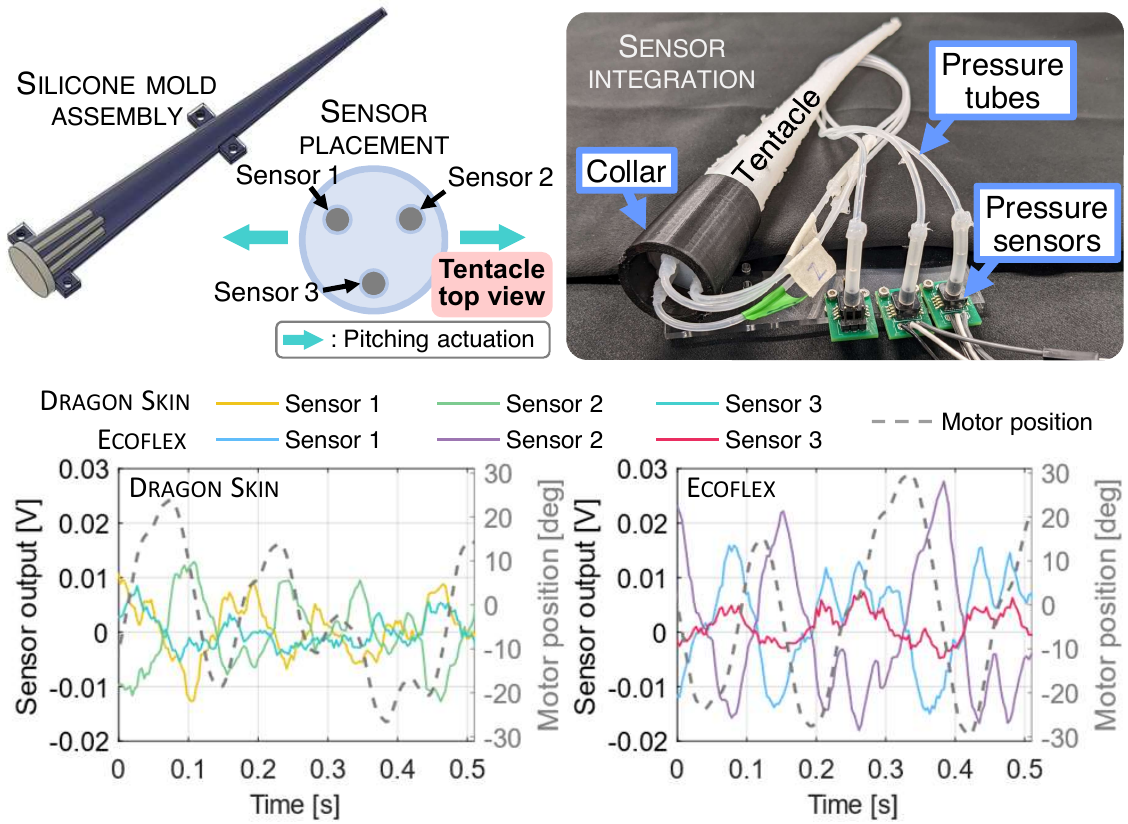}
    \caption{\textbf{Top}: Sensor design, fabrication, and pitching actuation direction for the sensorized tentacle. \textbf{Bottom}: Sample of raw sensor output from the tentacle actuated in a triangular wave motion with randomized amplitudes.}
    \label{fig:met_fab}
    \vspace{-2mm}
\end{figure}

The silicone tentacle is designed with voids at the root, which are attached to a pressure sensor (Fig.~\ref{fig:met_fab}). When the tentacle deforms due to the actuation at the base and the external forces from the water, the volume of the voids changes, which can then be used to proprioceptively infer the morphology of the tentacle. 
The tentacles are fabricated by 3D printing two-part molds and a cap with three protrusions used to create voids inside the tentacles. The tentacles have a length of $220\,\mathrm{mm}$ and a root diameter of $24\,\mathrm{mm}$. Two types of tentacles are created---a stiffer tentacle made of Dragon Skin\textsuperscript{TM} 10 and a softer one made of Ecoflex\textsuperscript{TM}---which enables comparison of various deformation profiles.

The voids in the tentacle are each connected to a MPXH6115A pressure sensor using flexible silicone tubes. A collar is mounted on the root of the tentacle to amplify the change in volume detected by the pressure sensor. The placement of the sensor holes is designed to capture the deformations of the structure when subject to a planar periodic motion at the root, which is constrained as a triangular wave with a frequency, $f$, and an amplitude, $A$.
Three sensors provide the optimal information content for reconstruction, as determined by reverse training a neural network, which is further discussed in Sec.~\ref{subsec:train_reconstruct}. 
A representative time series obtained from the pressure sensor while gathering training data is shown in Fig.~\ref{fig:met_fab} along with the position of the motor actuating the tentacle at the root. The patterns of the sensor output given the randomized actuation are challenging to analytically map, motivating the need for a training network. 


\subsection{Experimental setup \& data capture}

\begin{figure}[tb]
    \centering
    \includegraphics[width=1\columnwidth]{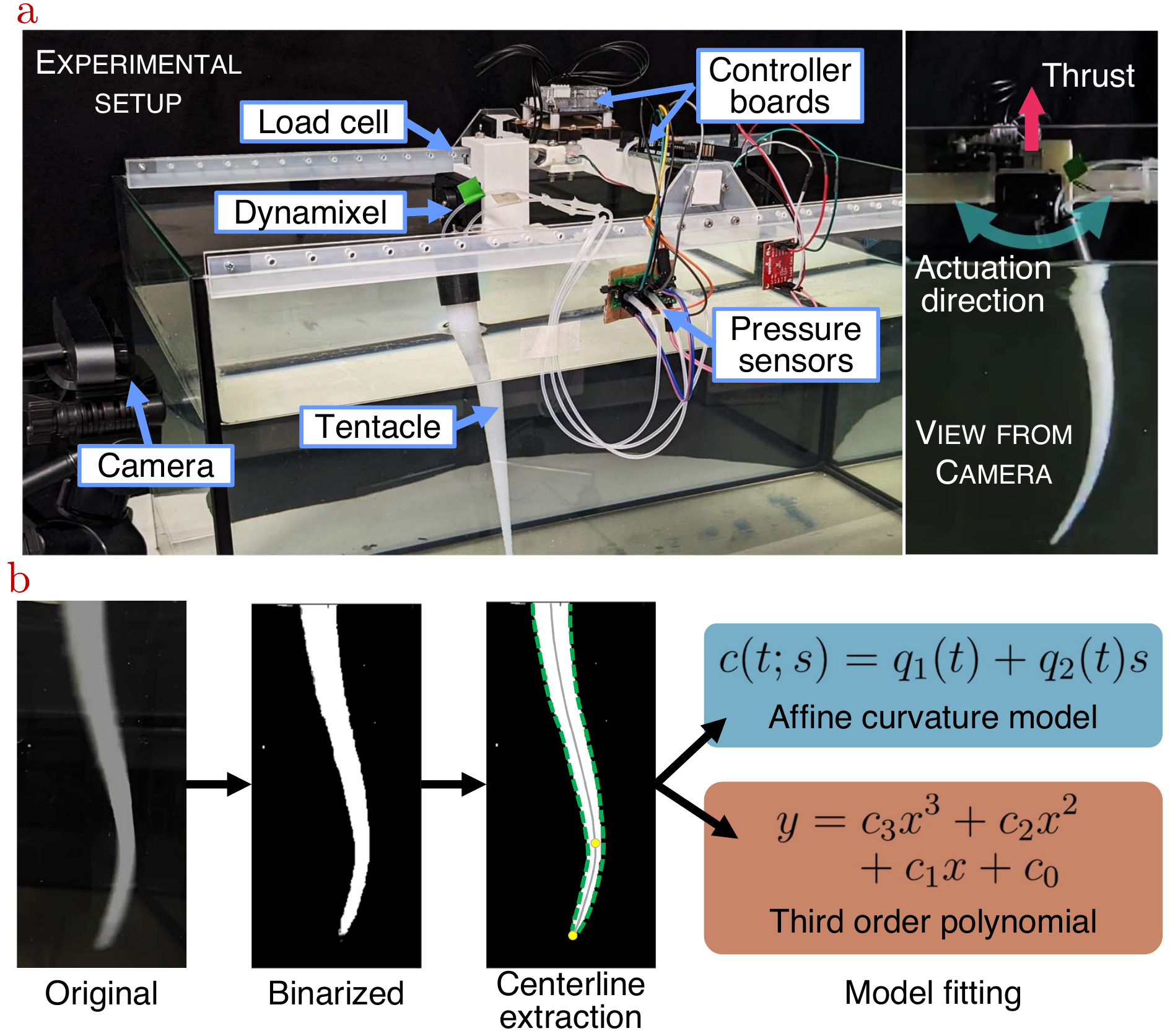}
    \caption{(\textbf{a}) Experimental setup of a single tentacle showing the direction of thrust generation as well as the pitching actuation. (\textbf{b}) Pipeline used to fit the models from the camera images. The image is first contrasted and binarized. Then, the contour of the tentacle is extracted and the middle line is computed as the mean of the two boundaries.}
    \label{fig:met_setup}
    \vspace{-4mm}
\end{figure}

To generate a training dataset and to explore sensor performance and proprioceptive metrics for a single tentacle, the experimental setup shown in Fig.~\ref{fig:met_setup}a is used to acquire the pressure sensor readings while the tentacle is actuated with a servo motor. Additionally, a $1\,\mathrm{kg}$ load cell is integrated to evaluate the correlation of the pressure data with upwards thrust. 
The training dataset of the sensor response over time is generated by capturing a 60 FPS video of the planar motion of the tentacle from a physical distance of $30\,\mathrm{cm}$.  This provides the ground truth of the tentacle pose.
Each frame of the video is binarized and the outline of the tentacle is identified to extract the centerline of the tentacle (Fig.~\ref{fig:met_setup}b), onto which a model is fit to represent the morphology of the tentacle. The raw backbone can be represented using different models, which confine the allowable configuration of the tentacle in different ways. 

\subsection{Model fitting of the tentacle deformation}

The tentacle centerline obtained from computer vision is converted into a useful representation that appropriately captures the tentacles curvature. We consider polynomial fitting and the affine curvature model~\cite{della2020soft}.

\subsubsection{Polynomial fitting}

A third-degree polynomial captures well most centerline shapes. The 2\textsuperscript{nd} and 3\textsuperscript{rd} degree coefficients, $c_2$ and $c_3$, represent the curvature of the tentacle centerline.  This provides a benchmark for assessing alternative models.

\subsubsection{Affine curvature model}
Experimental validations reported in~\cite{stella2022experimental} demonstrated that an affine curvature model can accurately describe the shape of slender soft elements when interacting with water making it suitable for representing a soft tentacle. Here we succinctly present the affine kinematics for this model~\cite{stella2022experimental}. We assume that the curvature $c(t;s) \in \mathbb{R}$ function of the central axis can be described by the affine function:
\begin{equation}
    c(t;s)=q_1(t)+q_2(t)s
\end{equation}
where $q = (q_1,q_2) \in \mathbb{R}^2$ are the Lagrangian coordinates of the state. The local spatial coordinate $s \in [0,1]$ parameterizes the position along the main axis of the tentacle. Here, $Ls$ is the arc length of the curve connecting the base to the point $s$ through the main axis. Note that--as common in the Cosserat framework--we assume the tentacle's cross-sections to be undeformable. Thus, the overall shape of the robot at each time $t$ is completely specified by the configuration of its central axis. The angle of the central axis $\alpha$ can be found by integration of the curvature: 
\begin{equation}
    \alpha(t;s)=\int_0^s c(\textcolor{black}{v}) \mathrm{d}\textcolor{black}{v} =q_1(t) s+ \frac{q_2(t)}{2} s^2,
\end{equation}
where $v$ is an auxiliary variable with the same meaning as $s$. The Cartesian position of point $s$ on the central axis is
\begin{equation}
    \begin{split}
        x_\mathrm{c}(t;s)&=-\int_0^{\textcolor{black}{s}} L \sin(\alpha\textcolor{black}{(t;v)}) \mathrm{d}\textcolor{black}{v}, \\ y_\mathrm{c}(t;s)&=+\int_0^{\textcolor{black}{s}}  L \cos(\alpha\textcolor{black}{(t;v)}) \mathrm{d}\textcolor{black}{v}, 
    \end{split}
\end{equation}
which admit closed form solutions in the form of Fresnel integrals~ \cite{della2020soft} and where $L$ is the undeformed length of the tentacle.
Fig.~\ref{fig:fig1} shows a schematic representation of the model of the tentacle, where the main kinematic quantities, $s$ and $\alpha$ are highlighted.


\subsection{Model training \& reconstruction of tentacle shape}
\label{subsec:train_reconstruct}

To reconstruct the tentacle profile from the sequence of raw sensor data, a bidirectional LSTM (long short-term memory) is used to train the network. An LSTM utilizes the past states to make predictions~\cite{siami2018comparison}. 
In the LSTM, stochastic gradient descent with momentum of 0.8 is used, the initial learning rate is set to 0.01 reducing by a factor of 0.85 every epoch, and there is a bidirectional LSTM layer with two fully connected layers. The network was trained with 35 epochs for the stiffer tentacle and 20 epochs for the softer one. 
The training data is constructed by obtaining the readings from the three pressure sensors and the polynomial fitting or affine curvature model obtained from computer vision. The tentacle is actuated at the root for 100 seconds with a randomized variable amplitude in the range $[-30,\ 30]^\circ$ each cycle with motor speed varying gradually from 12 to $80\,\mathrm{RPM}$. The test data is constructed similarly but lasts only for 40 seconds.



\begin{figure}[tb]
    \centering
    \includegraphics[width=1\columnwidth]{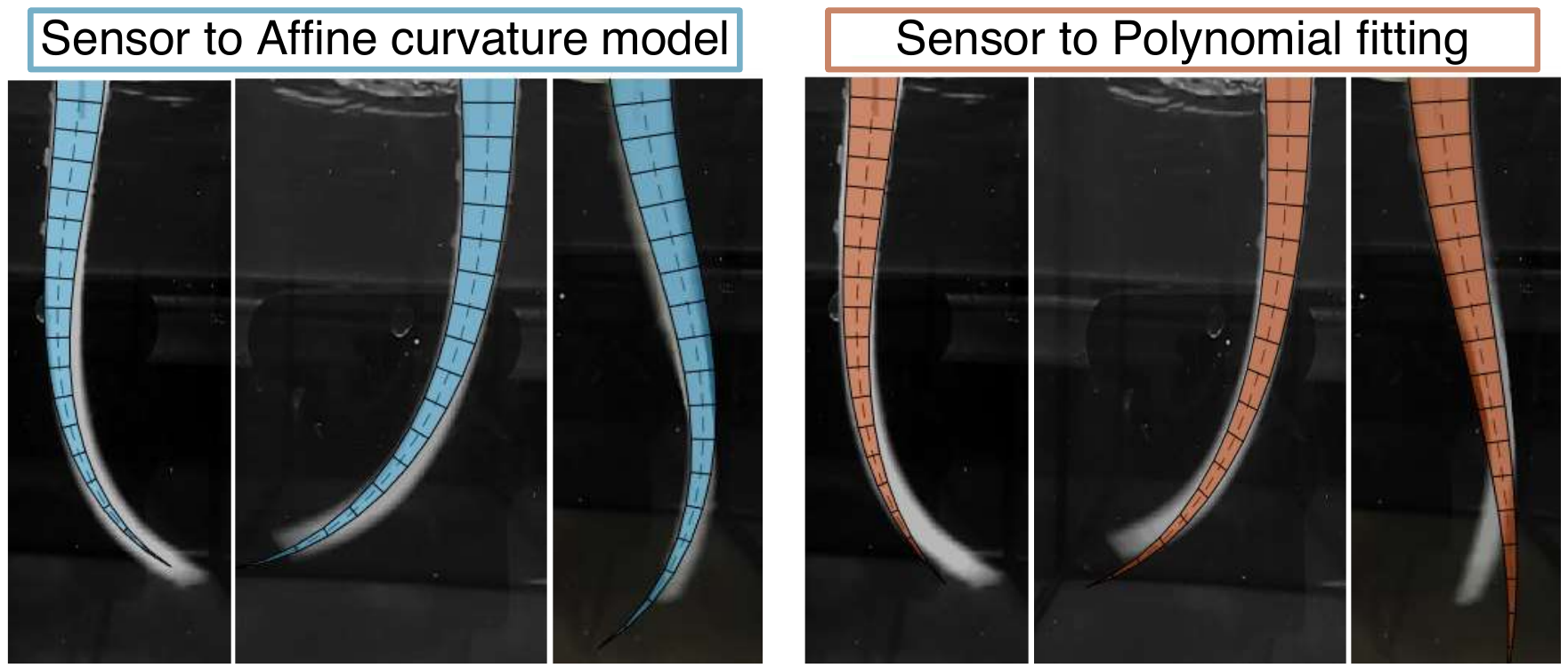}
    \caption{Trained reconstruction results from sensor readings overlaid with raw images using the affine curvature model (\textbf{Left}) and third order polynomial fitting (\textbf{Right}).}
    \label{fig:reconstruction}
    \vspace{-4mm}
\end{figure}

The reconstruction results for the polynomial and affine curvature fittings are shown in Fig.~\ref{fig:reconstruction}. The pressure data trained using the affine curvature model predicts the deformation better. 
The average error between the ground truth and the reconstruction using the test dataset for the two fitting methods is reported in Table~\ref{tab:error}. The normalized root-mean-square error (NRMSE) using the polynomial is calculated using the coefficients, $c_2$ and $c_3$. These coefficients describe approximately the deformation for segments 1 and 2, respectively (see Table key). The NRMSE using the affine curvature model is calculated using the resulting curvatures, $q_1$ and $q_2$ corresponding to segments 1 and 2, respectively. 
The relative tip error is calculated as a ratio of the absolute tip error to the maximum tip range for each tentacle. 
The reconstruction error for segment 2 (closer to the tip) is up to approximately 2\% lower than for segment 1. The affine curvature model performs especially well in reconstructing the tip. As we infer swimming performance from tip deflection in Sec.~\ref{subsec:metrics}, affine curvature-based reconstruction is used to obtain final results in this work. Furthermore, as indicated in the table by asterisks, the polynomial fitting fails to provide an accurate reconstruction of the tip for the Ecoflex tentacle which has low stiffness and considerably higher deformation. 

\setlength{\tabcolsep}{0.3em}
\begin{table}[tb]
\centering
\caption{Error in reconstruction when using different fitting models.}
\vspace{-2mm}
\label{tab:error}
\begin{tabular}{cccc}
 & Affine & Polynomial & Table key \\ \cline{2-3}
 & \multicolumn{2}{c}{\textbf{Dragon Skin}} & \multirow{10}{*}{
 \begin{minipage}{.14\textwidth}
      \includegraphics[width=\linewidth, height=43mm]{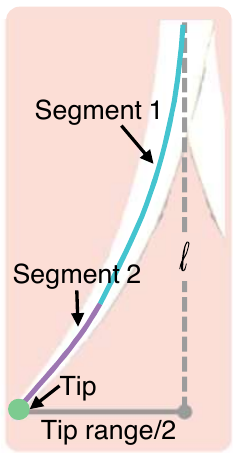}
    \end{minipage}} \\ \cline{1-3}
NRMSE\textsubscript{Seg1} & 7.8\% & 8.3\% &  \\
NRMSE\textsubscript{Seg2} & 6.6\% & 6.1\% &  \\
Abs. tip err. [mm] & 6.0$\pm$4.7 & 9.0$\pm$7.2 &  \\
Rel. tip err. & 4.6\% & 6.9\% &  \\ 
 & \multicolumn{2}{c}{\textbf{Ecoflex}} &  \\ \cline{1-3}
NRMSE\textsubscript{Seg1} & 5.7\% & 9.5\% &  \\
NRMSE\textsubscript{Seg2} & 5.5\% & 7.6\% &  \\
Abs. tip err. [mm] & 8.2$\pm$6.5 & 32$\pm$23* &  \\
Rel. tip err. & 5.4\% & 13.4\%* & 
\end{tabular}
\vspace{-4mm}
\end{table}

\subsection{Shape metrics for performance evaluation}
\label{subsec:metrics}

In this work, we investigate how a soft structure's proprioceptive sensing can be used to understand and robustly optimize its swimming performance. 
Soft swimmer deformations are an output of the fluid-structure interactions and are linked to the propulsive performances of the system. In addition to an increase in thrust with actuation frequency and amplitude, an increase in the tip deflection, calculated as $\delta_\mathrm{tip}=\tan^{-1}\left(\frac{\mathrm{tip\ range}}{2\ell}\right)$ from Tab.~\ref{tab:error} key, is an indicator of propulsion boost~\cite{quinn_scaling_2014}. 

Propagating waves in the deformation dynamics can occur due to the strong damping of the flow and improve the swimming efficiency~\cite{ramananarivo_propagating_2014}. Tapered geometries with thicker roots than tips, are found to favor traveling waves along the body while maintaining overall higher swimming capabilities~\cite{leroy2022tapered}. The traveling wave index (TWI) is a direct scalar metric obtained with the complex orthogonal decomposition~\cite{feeny_complex_2013}. The TWI quantifies how much of the total deformation is governed by a traveling wave pattern (TWI = 1) and how much by a standing wave pattern (TWI = 0).

To efficiently explore the control parameters and the deformation of the tentacle for performance metrics, Bayesian optimization is used with an objective to maximize the efficiency through the TWI. The TWI is selected as it reaches a peak when we increase the actuation frequency. We wish to maximize our data collection around this peak value.  By choosing a large exploration ratio of $0.8$, we allow the algorithm to widely explore while still focusing on the important peaks of the TWI. A single tentacle in the tank is actuated according to the chosen control parameters and the thrust and pressure data are collected for a specified number of periodic cycles. The temporal evolution of the tentacle centerline is reconstructed based on the pressure data, which is then used to determine the tip deflection and the TWI. 

\subsection{Robustness to the environment}

\begin{figure}[tb]
    \centering
    \includegraphics[width=1\columnwidth]{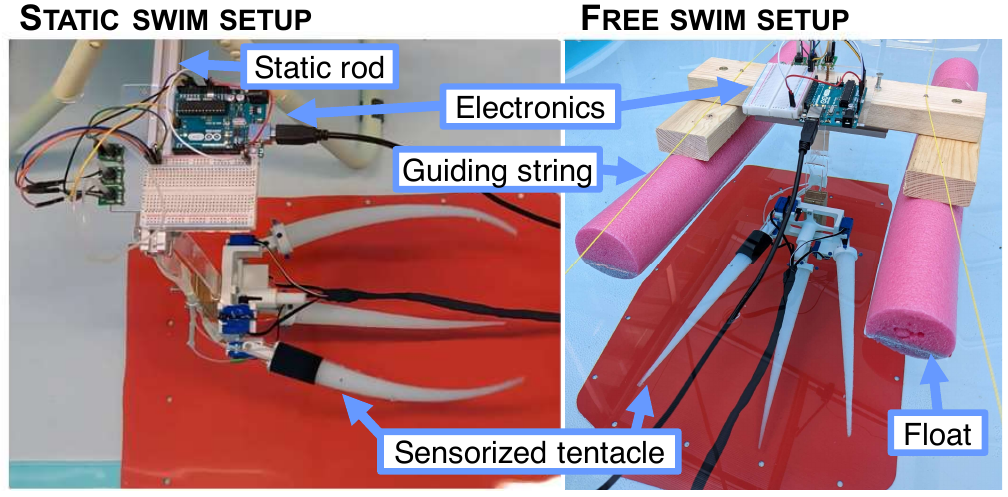}
    \caption{Full three-tentacle robot setups with labeled components used to perform static and free swimming tests.}
    \label{fig:met_swim}
    \vspace{-4mm}
\end{figure}

The embedded proprioceptive reconstruction allows the tentacle's control strategy to adapt to different environmental conditions, such as different flows of water, without the help of external sensors. 
To test this, the sensorized tentacle can be integrated into a robotic system along with two other unsensorized tentacles that are actuated simultaneously in the same control sequence. The robot swims both while rigidly attached and in a free swimming configuration (Fig.~\ref{fig:met_swim}) so that the sensor robustness can be evaluated.
In the rigid, or static swimming setup, the robot is mounted onto a static rod such that there is no flow, but computer vision can be used to obtain a ground truth of the tip deflection. In the free swimming setup, the robot swims along guiding strings with the help of floats to ensure the motion is constrained to a forward direction and is thus comparable between different controllers. In both setups, the robot is tethered to provide power and control signal.

\section{Results}

\subsection{Single tentacle proprioceptive sensing}

\begin{figure}[tb]
    \centering
    \includegraphics[width=1\columnwidth]{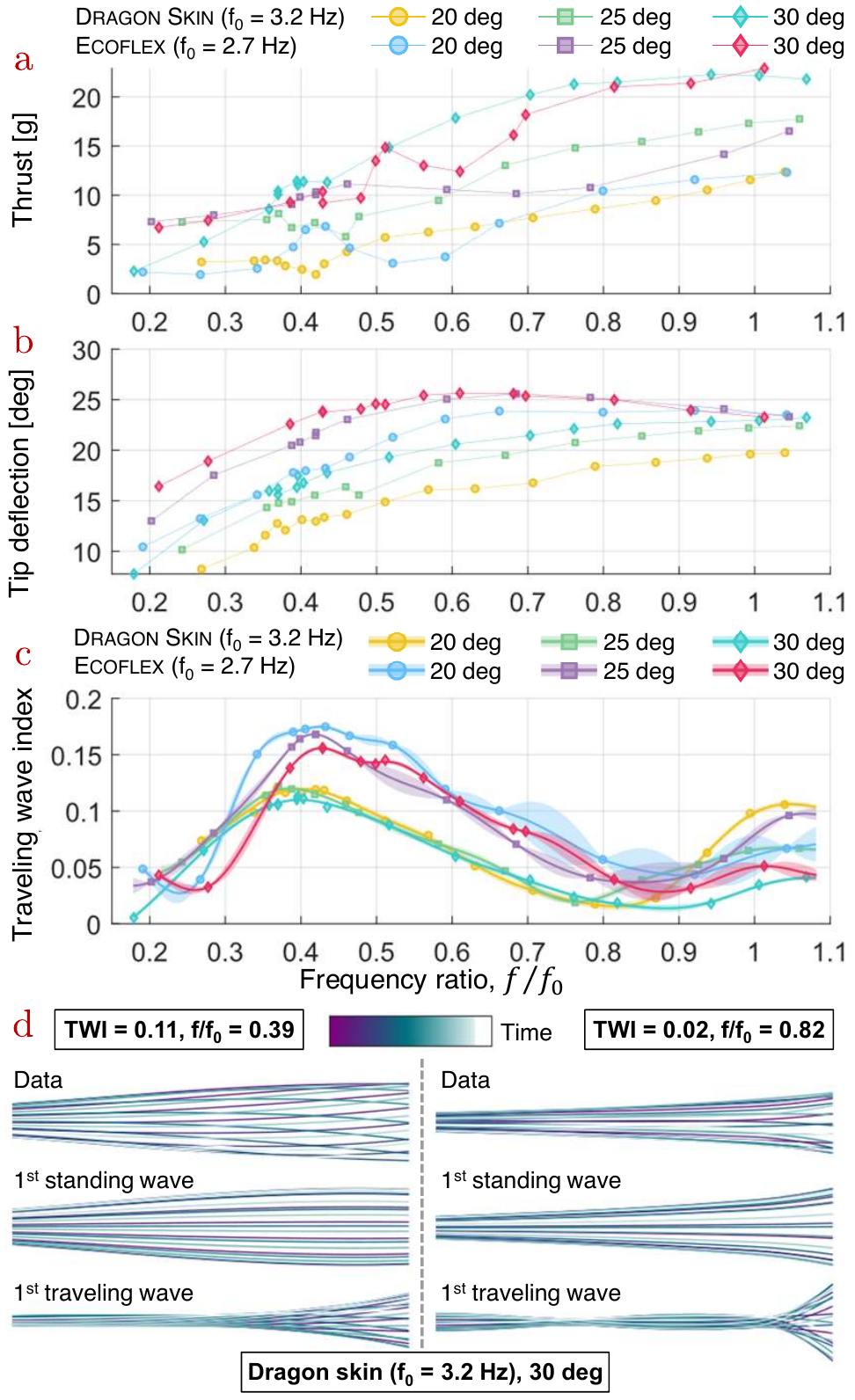}
    \caption{Thrust (\textbf{a}), tip deflection (\textbf{b}), and traveling wave index (\textbf{c}) for varying frequency ratios obtained through Bayesian optimization to maximize TWI. The 1\textsuperscript{st} standing and traveling waves from complex orthogonal decomposition are shown for two different TWI for comparison (\textbf{d}).}
    \label{fig:res_perf}
    \vspace{-4mm}
\end{figure}

Propulsive performances and shape metrics are obtained for a single tentacle for three different amplitudes at a range of frequencies (Fig.~\ref{fig:res_perf}). 
These are obtained via Bayesian optimization, with the objective to maximize the traveling wave index (TWI). The thrust is obtained from the load cell and the moving average with a window length, $k=3$ is shown to account for the large noise in load cell measurements due to the low absolute quantities of thrust. The TWI and tip deflection are reconstructed through the embedded sensor readings. To compare tentacles made of different silicones, the frequency is normalized by the natural bending frequency, $f_0$ in air found experimentally: $f_0=3.2\,\mathrm{Hz}$ for Dragon Skin\textsuperscript{TM} and $f_0=2.7\,\mathrm{Hz}$ for Ecoflex\textsuperscript{TM}.
An increase in thrust with respect to the driving frequency and amplitude is observed for both tentacles. The increase in thrust is related to an increase in tip deflection. The tip defection increases with increasing driving frequency and amplitude, but plateaus after $f/f_0 \sim 0.6$. The maximum tip deflection is reached at higher frequencies for lower driving amplitudes. This implies that the knowledge of the deformation can be used to tune the driving actuation in alternate environments when external sensing is not available.

A peak in traveling wave index is observed for both samples around the same frequency ratio $f/f_0 = 0.4$ (Fig.~\ref{fig:res_perf}c). The collapse of the curves for each tentacle suggests that the TWI does not depend on the driving amplitude. Snapshots of the first traveling mode reveal that the traveling wave amplitude is only significant in the tip region for the low TWI case whereas a larger portion of the body is affected for the high TWI case (Fig.~\ref{fig:res_perf}d).  Control of the frequency $f$ (or alternately of the stiffness that affects $f_0$) can be used to reach the TWI peak and increase propulsion efficiency. The results we obtain solely from the embedded sensor readings in the experimental setup for a single tentacle demonstrate a potential for using proprioceptive sensing and therefore its morphology to maximize or reliably control performance even when transferred to other environments.

\subsection{Sensor robustness validation with robot swimming}

To check how the reconstruction from the LSTM which is trained on data from the tank setup transfers to the full robotic system mounted horizontally in a pool, swimming experiments are performed with the full robot in the static and free swimming configurations (Fig.~\ref{fig:met_swim}). The robot has a total of three tentacles actuated simultaneously using the same controller, one of which is sensorized. The parameterized controller is offset $A/2$ degrees so that the interference between the tentacles is minimized. The servo motors used in the single tentacle experiments and the full robot experiments are different and so there may be variations in the motor output.

\subsubsection{Static swimming}

\begin{figure}[tb]
    \centering
    \includegraphics[width=1\columnwidth]{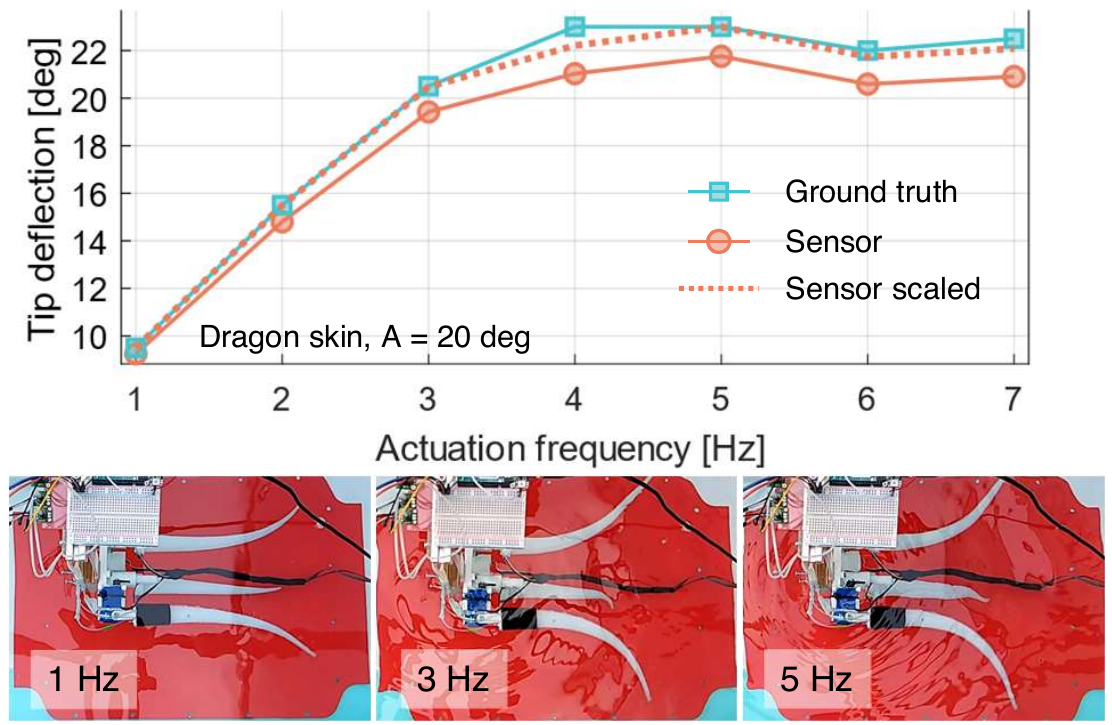}
    \caption{Reconstructed tip deflection from the sensor plotted with ground truth in the static swim test setup for the stiffer tentacle actuated with an amplitude of $20^\circ.$ Shots of the tentacle at maximum deflection are shown for chosen frequencies.}
    \label{fig:res_staticswim}
    \vspace{-4mm}
\end{figure}

Fig.~\ref{fig:res_staticswim} shows the tip deflection of the sensorized tentacle for a range of actuation frequencies $[1,\ 7]\,\mathrm{Hz}$ at an amplitude of $20^\circ$, while the robot is statically mounted. The predictions of tip deflection obtained from the reconstructed sensor readings are close to the ground truth results obtained from an overhead camera with a maximum error of only 8.6\%. To demonstrate the robustness of the embedded sensor reconstruction, the tip deflection from the sensor is scaled to match the range of the ground truth values and shown on the same plot, which shows excellent agreement. A photo of the robot exhibiting maximum deflection is shown for $f=1,\ 3,\ 5\,\mathrm{Hz}$ in Fig.~\ref{fig:res_staticswim}. The deformation profile varies significantly, with the tip deflection increasing with growing frequency and plateauing at $f\approx5\,\mathrm{Hz}$.  To understand how the reconstructed tip deflection of the tentacle on the robotic swimmer can be used to describe the performance of the swimmer and to validate the single tentacle experiments, it is necessary to quantify the thrust provided by the tentacles.

\subsubsection{Free swimming}

\begin{figure}[tb]
    \centering
    \includegraphics[width=1\columnwidth]{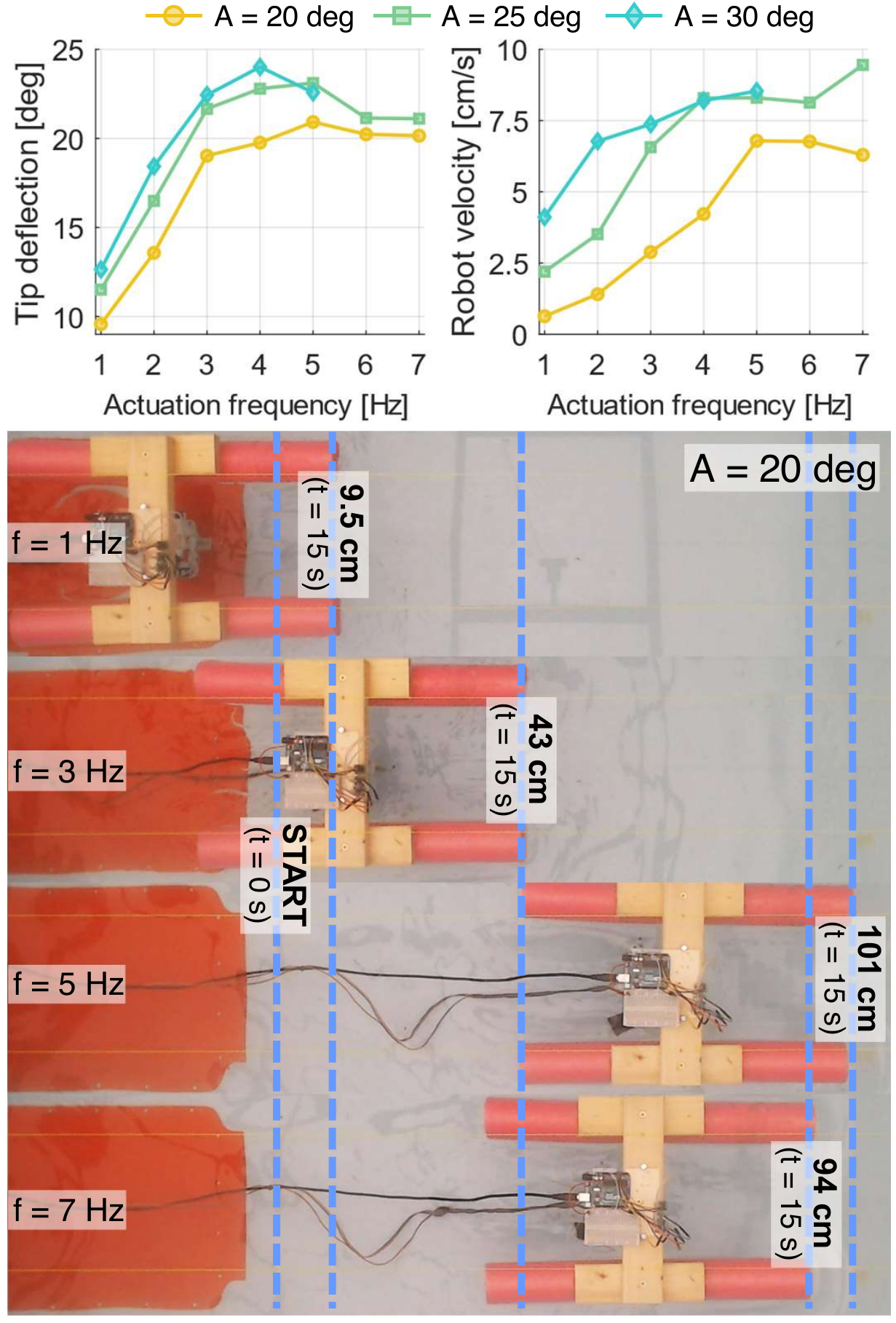}
    \caption{Reconstructed tip deflection and swimming velocities of the free swimming robot fitted with the stiffer tentacle. Swimming distance comparison is visualized for an actuation amplitude of $20^\circ$. For an amplitude of $30^\circ$, data is not presented beyond frequency of $5\,\mathrm{Hz}$ due to servo motor limitations in reaching the goal position.}
    \label{fig:res_freeswim}
    \vspace{-4mm}
\end{figure}

Fig.~\ref{fig:res_freeswim} shows the reconstructed tip deflection and the robot velocity in the free swimming setup for a range of actuation frequencies and amplitudes. The distance traveled by the robot for the actuation amplitude of $20^\circ$ is also shown. The robot's velocity increases with the actuation frequency and plateaus at $f\approx5\,\mathrm{Hz}$. The robot reaches a maximum swimming speed of approximately $7\,\mathrm{cm/s}$ traveling $101\,\mathrm{cm}$ in 15 seconds. At $f=7\,\mathrm{Hz}$, the robot velocity decreases and the robot only travels $94\,\mathrm{cm}$. 
These results validate the single tentacle reconstruction experiments and the correlation between tip deflection and thrust.
Since a clear correlation is observed between tip deflection (a shape characteristic) and velocity, this decrease in performance beyond a certain actuation frequency demonstrates the need for embedded sensors for proprioceptive information about its own morphology.

\section{Discussion \& Conclusion}
In this work, we explore the accuracy and robustness of an embedded proprioceptive sensor in a soft tentacle by inferring swimming performance from thrust, tip deflection, and traveling wave index. 
The pressure sensors at the base of the tentacle leave the body unhindered. The tentacle structure can be reconstructed based on the sensor data and a trained LSTM network. 
In the experiments, we demonstrate that the sensor readings contain the necessary information for accurate reconstruction in different usages and flow conditions. The characteristics of the thrust generated and the traveling wave index for varying actuation frequencies motivates the need for proprioceptive sensing without the aid of external sensors, as their relationship bewteen the control parameters are are not immediately discernible. We validated that tentacles with embedded sensors can `understand' and identify the tip deflection that leads to maximum thrust generation.  
This work motivates the inclusion of embedded proprioceptive sensors in soft structures. Further work should include the reconstruction in 3D, optimized sensor design, introduce external `objects' in the environment, and deploy the sensor to more complex soft structures to further leverage the morphological exploitation in underwater environments.


\addtolength{\textheight}{-12cm}   




\section*{Acknowledgment}
This project was partially supported by the EU's Horizon 2020 research and innovation program under the Marie Skłodowska Curie grant agreement N$^\circ$ 945363.

\bibliographystyle{IEEEtran}
\bibliography{references}

\end{document}